%% file: main.tex
\documentclass[runningheads]{llncs}

\usepackage{graphicx}
\usepackage{booktabs}
\usepackage{multirow}
\usepackage{amsmath,amssymb}
\usepackage{fancyvrb}
\usepackage{verbatim}
\usepackage[most]{tcolorbox}
\usepackage{enumitem,enumerate}
\usepackage{color,soul}
\usepackage{times,latexsym}
\usepackage{url}
\usepackage[T1]{fontenc}
\usepackage{graphicx}
\usepackage{booktabs}
\usepackage{multirow}
\usepackage{amsmath,amssymb}
\usepackage{fancyvrb}
\usepackage{verbatim}
\usepackage[most]{tcolorbox}
\usepackage{enumitem,enumerate}
\usepackage{color,soul}
\usepackage{makecell}
\usepackage{soul}

\usepackage[utf8]{inputenc}

\usepackage{xargs}  
\usepackage[colorinlistoftodos,prependcaption,textsize=tiny]{todonotes}

\newcommandx{\unsure}[2][1=]{\todo[linecolor=red,backgroundcolor=red!25,bordercolor=red,#1]{#2}}
\newcommandx{\change}[2][1=]{\todo[linecolor=blue,backgroundcolor=blue!25,bordercolor=blue,#1]{#2}}
\newcommandx{\info}[2][1=]{\todo[linecolor=OliveGreen,backgroundcolor=OliveGreen!25,bordercolor=OliveGreen,#1]{#2}}
\newcommandx{\improvement}[2][1=]{\todo[linecolor=Plum,backgroundcolor=Plum!25,bordercolor=Plum,#1]{#2}}
\newcommandx{\thiswillnotshow}[2][1=]{\todo[disable,#1]{#2}}
\newcommand{\modelname}[1]{{\small \texttt{#1}}}
\newcommand{\quotedexample}[1]{{\emph{"#1"}}\normalsize}

\newcommandx{\revision}[1]{\textcolor{black}{#1}}
\newcommandx{\tocheck}[1]{\textcolor{black}{#1}}

\author{Besnik Fetahu \and
Nachshon Cohen \and
Elad Haramaty \and
Liane Lewin-Eytan \and
Oleg Rokhlenko \and
Shervin Malmasi}
\authorrunning{Fetahu et al.}

\institute{
Amazon.com, Inc. ~~~ Seattle, WA, USA \\
\email{\{besnikf,nachshon,eladh,lliane,olegro,malmasi\}@amazon.com} 
}

\title{Identifying Shopping Intent in Product QA for Proactive Recommendations}

\begin{document}
\maketitle

\begin{abstract}
Voice assistants have become ubiquitous in smart devices allowing users to instantly access information via voice questions. While extensive research has been conducted in question answering for voice search, little attention has been paid on how to enable proactive recommendations from a voice assistant to its users. This is a highly challenging problem that often leads to user friction, mainly due to recommendations provided to the users at the wrong time.
We focus on the domain of e-commerce, namely in identifying \emph{Shopping Product Questions} (SPQs), where the user asking a product-related question may have an underlying shopping need. Identifying a user's shopping need allows voice assistants to enhance  shopping experience by determining \emph{when} to provide recommendations, such as \emph{product} or \emph{deal} recommendations,  or \emph{proactive shopping actions recommendation}.
Identifying SPQs is a challenging problem and cannot be done from question text alone, and thus requires to infer latent user behavior patterns inferred from user's past shopping history. We propose features that capture the user's latent shopping behavior from their purchase history, and combine them using a novel Mixture-of-Experts (MoE) model. 
\revision{Our evaluation shows that the proposed approach is able to identify SPQs with a high score of F1=0.91. Furthermore, based on an \emph{online} evaluation with real voice assistant users, we identify SPQs in real-time and recommend shopping actions to users to add the queried product into their shopping list. We demonstrate that we are able to accurately identify SPQs, as indicated by the significantly higher rate of added products to users' shopping lists when being prompted after SPQs vs random PQs.}

\end{abstract}

\input{introduction}
\input{related_work}

\input{datasets}

\input{approach}

\input{setup}
\input{conclusions}

\bibliographystyle{splncs04}
\bibliography{references}

\end{document}

%% file: introduction.tex
\section{Introduction}\label{sec:introduction}

Voice assistants, like Alexa or Google Assistant provide ubiquitous services through a variety of devices (e.g. smart speakers, phones, TVs etc.). Users interact with voice-assistants for different  purposes~\cite{rzepka2019examining,lopatovska2019talk} such as question answering \cite{faustini-etal-2023-answering}, task completion \cite{choi-etal-2022-wizard}, conversational search \cite{salle2021studying}, entertainment, or control of smart devices. Determining the underlying \emph{user intent} is an active field of research ~\cite{DBLP:conf/icassp/TyagiSGSZWC20,le-etal-2021-combining,DBLP:conf/www/YangDDM18}, given that new skills are continuously added to such voice assistants. 

In this context, making proactive follow-on suggestions is an active area of research \cite{fetahu2023followon}. However, key challenges with voice-based conversational recommender systems remain, such as their failure to adapt to evolving user  behavior~\cite{DBLP:journals/corr/abs-1901-00431}. Additionally, interactions are typically initiated by users, and proactive system recommendations may lead to increased user friction. Knowing \emph{when} to proactively recommend actions or items to users on the next turn, such as suggesting the right product recommendations~\cite{10.1145/3523227.3546761,DBLP:conf/recsys/MaLYYZ22,DBLP:conf/sigir/MengYX20,DBLP:conf/cikm/HidasiK18,DBLP:conf/cikm/LiRCRLM17,DBLP:conf/kdd/LiuZMZ18} or next actions, is tightly dependent on accurately identifying the user's underlying intent. Correctly identifying this intent can avoid user dissatisfaction by providing recommendations only when necessary.
Our work focuses on identifying the right time to make a proactive personalized recommendation in a voice-based conversational system.

\begin{figure}[ht!]
    \centering
    \includegraphics[width=0.9\columnwidth]{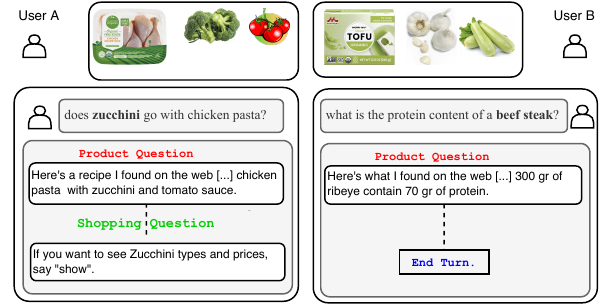}
    \caption{\small{User A searches for ingredients for a \emph{recipe}. Our approach identifies the user having a shopping need, hence, triggering voice assistant's shopping recommendations. For User B's (likely vegetarian) our model predicts no shopping intent, hence, after answering, the turn is over.}}
    \label{fig:spq_conv_example}
\end{figure}

E-commerce is an important functionality in voice assistants. In some contexts, the shopping intent can be explicit, such as in product searches (e.g. \quotedexample{Buy me dog food}, or \quotedexample{Search for an Apple watch stand}). However, intent cannot always be captured from the question alone, e.g., product-related questions, such as: 
\quotedexample{What can I use as a substitute for toilet paper?}; \quotedexample{How long do you cook rice?}; \quotedexample{Are dogs allowed to eat tuna?}.

The examples show that the user intent is often not clear from the Product Question (PQ) itself. PQs may emerge from an underlying desire to potentially purchase a product (\emph{shopping need}), or only for a \emph{general} knowledge need.

Voice product search~\cite{DBLP:conf/wsdm/CarmelHLLM20,DBLP:journals/chb/RheeC20} and question answering~\cite{DBLP:conf/sigir/Guy16,carmel2018product,DBLP:conf/eacl/RavichanderDRMH21,DBLP:conf/naacl/RamnathSHY21} have received attention, while understanding the reasons why users inquire information about certain products, especially in terms of \emph{shopping need}, has seen very little progress.
This is the main challenge that we tackle, namely determine whether a user has a shopping need. This allows voice assistants to determine \emph{when} to surface recommendations that improve user's shopping experience, e.g. after answering a product question, show product recommendations, deals or promo suggestions, or offering additional product details for further examination.  In this work, we do not focus on \emph{what} to recommend given the vast literature in product (sequential) recommendation~\cite{10.1145/3523227.3546761,DBLP:conf/recsys/MaLYYZ22,DBLP:conf/sigir/MengYX20,DBLP:conf/cikm/HidasiK18,DBLP:conf/cikm/LiRCRLM17,DBLP:conf/kdd/LiuZMZ18}.

Figure~\ref{fig:spq_conv_example} shows a hypothetical user interaction with a voice assistant. Depending on the user's interests, shopping recommendations may differ. For instance, for \emph{User A}, based on their purchase history we can infer that they may be interested in purchasing a queried product, recommending shopping related actions, and thus ease the shopping process for the user. For \emph{User B} on the other hand, who has vegetarian preferences, the intent is general knowledge need.

Identifying Shopping Product Questions (SPQ) in voice search poses several challenges. Contrary to shopping intent prediction in e-commerce sites~\cite{esmeli2020towards,ling2019customer,DBLP:conf/kdd/GuoHJZW019}, signals such as click-rate, browse time, hardware related gestures are not present. Furthermore, shopping need detection in voice search is inherently harder given the multi-purpose use of voice assistants, compared to a restricted use of e-commerce sites.

We propose an approach based on Graph Attention Networks (GAT)~\cite{DBLP:conf/iclr/VelickovicCCRLB18} to identify shopping need. Questions are considered nodes and are connected if they share the same product. As input features, we consider features that are geared at capturing diverse and latent aspects of shopping need, such as \emph{product information}, user \emph{past purchasing behavior}, and the \emph{question} itself. Finally, based on the mixture-of-experts approach (MoE)~\cite{DBLP:conf/iclr/ShazeerMMDLHD17}, we propose a mechanism to compute a joint node representation from the diverse features.

\revision{Experimental evaluation on more than 370k voice assistant user questions shows that our proposed approach allows us to achieve highly accurate results in distinguishing SPQs. Furthermore, we carry out an online experiment with real voice assistant users, where we identify SPQs in real-time, and recommend users to add the queried product into their \emph{shopping lists} following a SPQ, identified with high accuracy.}\footnote{As future work we plan to experiment with different types of recommended shopping actions and assess the impact on the user experience.}
\tocheck{The approach presented in this paper through detection of SPQs allows voice assistants to recommend personalized responses according to user's interests and needs, such as {``\emph{check price''}}, {\emph{``show deals''}}, {\emph{``add to cart''}}, {\emph{``compare products''}} etc.}
Our main contributions in this work are as follows:
\begin{itemize}[leftmargin=*]
    \item \revision{A new problem definition for identifying SPQs that enable voice assistants to determine when to recommend to users.} %
    \item \revision{An approach to identify SPQs in voice search, using a novel way to combine diverse features through Mixture of Experts stemming from various user signals and containing different feature types.}
    \item As part of these experiments, we perform detailed ablations showing the impact of the different features in identifying queries with shopping need, highlighting the latent nature of users shopping behavior, and thus opening directions for improving users shopping experience with voice assistants.
    \item  \revision{Detailed offline and online experimental evaluation on voice assistant users and user queries. }
    
\end{itemize}

%% file: related_work.tex
\section{Related Work}\label{sec:related_work}

\textbf{Users Needs in E-commerce.} 
Informational need in e-commerce and voice shopping is typically expressed by asking questions about products. This domain is also referred to as Product Question Answering (PQA), and has attracted much interest in the past years ~\cite{carmel2018product,Chen:2019,Kulkarni:2019,zhang:2019,gao+al:19,rozen+al:21}. Some research in the PQA domain has also been conducted with respect to the voice medium \cite{DBLP:conf/sigir/Guy16,carmel2018product,DBLP:conf/eacl/RavichanderDRMH21,DBLP:conf/naacl/RamnathSHY21}, where queries have different characteristics, and where cross-lingual and speech-to-text components add another layer of complexity.
Most of the works in the PQA domain focus on different answering approaches and on different types of questions \cite{rozen+al:21,xu:18,mcauley+Yang:16,yu+al:12,lai+al:18,lai+al:18b}. Here, we do not focus on the answering side at all, but rather focus on the informational need behind product questions.

Transactional need in e-commerce providers has been investigated, mainly through purchase intent prediction ~\cite{esmeli2020towards,ling2019customer,DBLP:conf/kdd/GuoHJZW019,DBLP:conf/sigir/AhmadvandKJA20,DBLP:conf/www/ShenYYJLC11}, where behavior features correspond to browse or hardware related (touch interaction on smartphones)~\cite{DBLP:conf/kdd/GuoHJZW019}. Esmeli et al.~\cite{esmeli2020towards} extract features such as product click rate, view time, number of visits, to predict purchase intent. Our key differences with these works is that we tackle the problem of voice modality, where voice assistants are used for highly diverse purposes~\cite{lopatovska2019talk,rzepka2019examining}, which is not the case of dedicated e-commerce sites, where the primary purpose is for shopping. Secondly, features that can be extracted from Web sites are not available for voice assistants, making the problem of predicting shopping needs of voice users more challenging. The work of \cite{haramati2020} explores the problem of predicting the purchase rate of a ranking model in a voice setting, where there is an extreme position bias towards the first offer. 
This work lies in the domain of voice product search, where the user's intent is clearly transactional -- unlike our work, which aims to identify an underlying shopping need in the voice PQA domain.

\textbf{Product Search and Recommendations.} There has been a large body of research in product search and product recommendations~\cite{DBLP:conf/sigir/BiAC21,10.1145/3437963.3441754,10.1145/3132847.3133060,10.1145/3018661.3018665,DBLP:conf/recsys/WangSS11,DBLP:conf/cikm/ZhaoCY19}, which are crucial in the shopping domain, for providing relevant products, and recommendations reflecting users' preferences. The voice medium has also attracted research in this context \cite{DBLP:conf/wsdm/CarmelHLLM20,DBLP:journals/chb/RheeC20}, as shopping queries and behavioral shopping patterns in voice are different than in web \cite{ingber:2018}. Learning product embeddings is another challenge that has received much attention ~\cite{10.1145/3340531.3412732,DBLP:conf/recsys/TagliabueYB20,DBLP:conf/recsys/VasileSC16}, as having relevant product embeddings allows for more accurate recommendations and better user experience. Our goal is different in that, the product of interest is explicit in the user query. Our goal is to {\em understand} the underlying user's intent, rather than to provide the most relevant recommendation.  Nevertheless, advanced product embeddings used for product recommendation might help in capturing the customer affinity to a specific product and  improve further the shopping need classification. 

\textbf{Behavioral Features in Search and Recommendation.} The works in \cite{DBLP:conf/sigir/AgichteinBD06,DBLP:conf/kdd/Joachims02} propose the integration of implicit user feedback for ranking Web search results. Although these works tackle a different problem and are not comparable to our work, we share a similar goal in that we obtain user behavior features (e.g. purchase history) that can serve to learn a latent user shopping intent behavioral model.

Works on session based recommendation~\cite{DBLP:conf/sigir/MengYX20,DBLP:conf/cikm/HidasiK18,DBLP:conf/cikm/LiRCRLM17,DBLP:conf/kdd/LiuZMZ18} propose approaches for incorporating diverse features to facilitate recommendation within a given user session (e.g. music search). At its core it predicts the next item with which the user will interact. These works are complementary to ours as they address the question on \emph{what} to recommend, whereas we focus on determining \emph{when} to trigger recommendations by the voice assistants recommender systems.

%% file: datasets.tex
\section{Data Analysis}\label{sec:datasets}
\subsection{Dataset and Labels} 
Since we cannot observe the true user intent, we rely on their purchase behavior: if the user asked a question and then purchased the referred product, we consider it as having a  shopping need.

A shopping intent does not often result in an instantaneous purchase.  
Selecting the time window for the purchase information, from the time the user issued a PQ up to the actual purchase, represents an interesting tradeoff. Longer periods tend to better capture the hidden purchase intent, however at the same time pose the risk of labeling  unrelated utterances with shopping intent. We analyzed different time windows, including one hour/day/week/two weeks, and 28 days, and found that a period of 28 days increased coverage by $187\%$ while increasing noise by $8\%$ only\footnote{We estimated noise by scrambling the utterances and purchases and measure how often an utterance is coincidentally followed by a purchase.} compared to a period of one week.
Given a PQ, we checked the future user purchases. 
If the referred product -- or a similar product, sharing the same category -- was purchased within 28 days, we label the PQ as SPQ, otherwise NSPQ.

We collected a dataset of 374k PQs from a leading voice assistant, evenly distributed among SPQs and NSPQs\footnote{
\revision{The sample was designed to have an even distribution of NSPQs and SPQs, as such it is not representative of the overall traffic and none of the results reported here should be extrapolated to the entire traffic}}, split into 306k/34k/37k subsets for train/validation/test set, respectively. For each question, we collected its text, voice assistant's response, the product and its category, and user past purchases.

\subsection{Data Insights}
To understand correlated signals with SPQs, we perform a correlation analysis (Pearson's correlation coefficient), between the question type, their timing, and the searched product.

\subsubsection{Questions.}
\begin{figure}[h!]
    \centering
    \includegraphics[width=0.60\columnwidth]{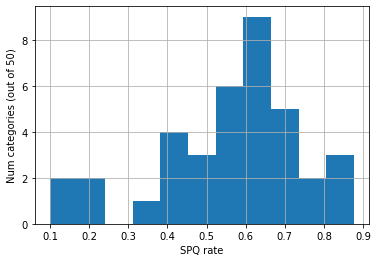}\vspace{-10pt}
    \caption{\small{SPQ rate per product type (out of 50 types with at least 1000 PQs, which represent 29.7\% of the entire dataset).}}
    \label{fig:spq_product_type_hist}
\end{figure}

The relation between the user's shopping need and the question text is involved. 
We found that questions asked from users about vitamins or pet-food have a high shopping intent (88\%, 82\%), while questions about televisions or cellphones have low shopping intent (only 16\%, 18\%). 
Figure~\ref{fig:spq_product_type_hist} shows the histogram of the SPQ rate for the 50 most popular product categories. %
We see that different product types have different SPQ rates compared to the average 50\%. 
There are product types where most of the queries are general knowledge need (4 categories with SPQ rate below 0.25\%), while there are product types where most questions have a shopping need (5 categories with SPQ rate above 0.75\%). 
For the rest, more than half of the popular categories have SPQ rate above 60\%. 
This demonstrates that the product category, which can be inferred from the text of the question, is indicative of SPQs.

However, while a shopping need is correlated with some textual features of the questions, this correlation is not natural nor intuitive. 
To demonstrate this, we sampled 10K questions and presented them to annotators. 
The annotator was presented with the question, the voice assistant response, and was required to estimate whether the user will likely purchase the referred product. 
To demonstrate this, we labelled 10K utterances, asking annotators to predict whether the user will purchase the product. 
We saw a very low correlation of 0.05 between human evaluation and future purchases.
This indicates that from text alone, humans cannot predict whether a question has a shopping intent. 
An example of this phenomena, e.g., \quotedexample{``what do fish eat?''}, which to annotators seems like a general knowledge need question, while later purchases of fishing materials demonstrate the latent shopping need behind this question.

\subsubsection{Users.} 
The user properties also have an involved relation with the shopping intent. 
On the one hand, we found that shopping need is not a user property. 
We took two random PQs from each user, and measured the correlation between the first one being an SPQ and the second being a NSPQ. 
To avoid two questions referring to the same product category, which could lead to a false correlation, we de-duplicated PQs from the same user about the same product category. 
The correlation was only $r=0.14$%
, showing that the same user may ask questions that have or do not have a shopping need.

On the other hand, we found a relation between the user's shopping history and the shopping need. 
For each PQ, we checked if the referred product was previously purchased by the user (28 days prior). 
This property has a strong correlation of $r=0.67$ with the SPQ property, which is based on a future purchase. 
This shows that when the user asks about a product category that was previously purchased, they are more likely to purchase again. 

While there is strong correlation between SPQs and preceding purchases, this might simply reflect user tendency to repurchase products, and be unrelated to the question. 
To refute this possibility, we measured the correlation of purchasing the referred product during two consecutive 28-days periods, before the actual user's question (PQ). We found that the correlation is significantly weaker when compared to consecutive periods coming before and after the user's PQ, showing that the PQ plays an important role. More specifically, given a PQ referring to product $P$ from the user's shopping history, we consider four 28-days time periods $T_0,T_{-1},T_{-2}$ and $T_{-3}$, where $T_0$ is the time period coming after asking the question, $T_{-1}$ is the time period coming before the question, and consecutively $T_{-3} < T_{-2} < T_{-1}$. The correlation between purchasing product $P$ during $T_0$ and purchasing $P$ during $T_{-1}$ is measured to be $r=0.67$. If this correlation was due to the periodic nature of purchasing $P$, we would expect a similar correlation of purchasing $P$ also during $T_{-2}$ and  $T_{-3}$. However, the correlation was $r=0.18$, showing that the high correlation of purchasing $P$ between times $T_0$ and $T_{-1}$ is strongly connected to the PQ, which indicates its shopping intent.

%% file: approach.tex
\section{Identifying SPQs}
\label{sec:approach}

To identify SPQs we present a model \modelname{SPQI} with three components: (a) features of different types, (b) feature aggregation through mixture of experts~\cite{DBLP:conf/iclr/ShazeerMMDLHD17}, and (c) a Graph Attention Network.

\noindent\textbf{Textual Features.} The PQ is an important intent indicator, containing key aspects that are asked about a product (e.g., \emph{price}, \emph{delivery time}, etc.). Additionally, voice assistant's response to the question contains information if the question was understood and answered.
Furthermore, when a customer asks a question the voice assistant responds with either an \emph{answer} or indicates that it \emph{does not understand} the utterance. 
For cases where the voice assistant provides an answer, the user can continue their conversation by either asking additional questions regarding the product or make a shopping decision. 
We encode textual features extracted with a RoBERTa model~\cite{DBLP:journals/corr/abs-1907-11692}, and use the \texttt{[CLS]} token for representation.

\noindent\textbf{Product Features.} An important aspect of a PQ is the queried product itself.
Products have a rich structure, and users may have shopping preferences towards seemly unrelated products or specific categories, e.g., because they share a similar price range or a similar theme, e.g., \emph{``zucchini''} and \emph{``broccoli''}, or \emph{``healthy food''}.
We use the product and the product category embeddings\footnote{Products are organized in a taxonomic product category graph} that are trained from scratch to capture shopping need related features.

We compute the product $p$, product category $c$, and parent category $\hat{c}$ of $c$ embeddings.
The features are considered independently from each other\footnote{As future work we foresee integrating graph embeddings for categories and products.} and allow the model to learn how to leverage them during training.

\noindent\textbf{Behavioral Features.} To identify shopping intent, the user's shopping behavioral patterns, extracted from the purchase history, are key. The purchase history is a sparse feature, consisting of the tuples $\langle$user, product$\rangle$. 
We pre-train a skip-gram based embeddings~\cite{DBLP:journals/tacl/BojanowskiGJM17} for better generalization. We use the tuple $\langle u, p\rangle$ to train a model that predicts if the product $p$ will be purchased by user $u$, where $u, p \in \mathbb{R}^{50}$ represent the product and user vectors. The resulting embedding have two highly desirable properties: (1) create user vectors that are similar according to purchasing patterns, and (2) product vectors that are similar according to the users that co-purchased them. 

Using the pre-trained model, for an input question, from the list of purchases made by the user in previous 28 days $\mathbf{H}$, we compute: (i) average dot/cosine similarity, maximum dot/cosine similarity, and the sum of dot/cosine similarity.

\subsection{Learning Joint Question Representations via MoE}
An important question is how to combine the various feature types (textual, numerical, categorical). Typically, such features are combined using either max or average pooling mechanisms~\cite{DBLP:conf/sigir/GaoLXWKL20,DBLP:conf/mm/WeiWN0HC19}. One drawback of this is that they do not allow the models to learn weights for each feature type according to their impact on the classification task. 

We propose the use of mixture-of-experts~\cite{DBLP:conf/iclr/ShazeerMMDLHD17}, which allows the dynamic mixing of the different question feature types. Let $\mathbf{F} \in \mathbb{R}^{f\times n}$ be the concatenated set of features. For each feature dimension in $\mathbf{F}$, MoE weighs the importance of the $f$ individual feature types as follows.
\small{\begin{align}
        \texttt{MoE}= \mathbf{W}'_e\left(\mathbf{F}\mathbf{W}_e + c_1\right)^{T} + c_2; \;\;\;\;\;\;
    \lambda_{f_j,z} = \frac{\text{\texttt{MoE}}_{f_j,z}}{\sum_{f'\in |f|}\text{\texttt{MoE}}_{f',z}}
\end{align}}\normalsize
where $\mathbf{W}_e\in \mathbb{R}^{n\times i}$, $\mathbf{W}'_e\in\mathbb{R}^{n\times i}$, represent trainable parameters used to compute feature type weights. 
Finally, $\lambda_{f_j,z} \in \mathbf{\Lambda}$ represents the feature weight for dimension $z$ and feature type $f_j$, 
while $\mathbf{\Lambda}\in\mathbb{R}^{f\times n}$ represents all the feature weights. %

The final joint question representation that is passed onto the GAT framework, is computed by performing the Hadamard product between each feature vector and the computed MoE weights: $\mathbf{h}=\sum_{f'\in |f|}\mathcal{F}_{f'}\circ \mathbf{\Lambda_{f'}}$.

\subsection{Question Graph Construction \& Classification}\label{subsec:spq_classification}

Considering questions as independent instances has the disadvantage of not being able to leverage similar \emph{questions} or \emph{users} that may have similar purchasing behavior.
Hence, we formulate the SPQ identification problem as a node classification task using GAT~\cite{DBLP:conf/iclr/VelickovicCCRLB18}. GAT allow to create soft dependencies between questions (if they share the same product), and as such, influences the prediction of shopping need for a question, by taking into account questions about the same product from other users.
An important aspect of utilizing GATs, is the shape of the graph. As nodes are the user questions, represented by the joint representation $\mathbf{h}$, and nodes are connected if they share the same queried \emph{product}.

We train a GAT model to compute the question representations: $\mathbf{\widehat{h}}_q=\text{\texttt{GAT}}(\mathbf{h}_q)$, which is used to obtain the SPQ probability as $\mathbf{s}_{q}=\sigma(\mathbf{\widehat{h}_{q}} \cdot \theta)$ to classify into either an SPQ or a NSPQ.

%% file: setup.tex
\section{Experiments}\label{sec:setup}

\revision{We discuss the offline and online evaluation results, and further introduce baselines for the offline experimentation phase.}

\subsection{Baselines and Ablations}
\noindent\textbf{RoBERTa-Baseline.} This baseline highlights the difficulty of identifying shopping intent from text alone. We consider two variations: (1) \modelname{RoBERTa-Question}, where only the question text is used for training and inference, and (2) \modelname{RoBERTa-Text}, where additionally voice assistant's answer is used for training and inference.

\noindent\textbf{MLP-Baseline.} All the computed features are pushed through an MLP layer for classification. We consider several ablations (\modelname{MLP-Question} and \modelname{MLP-Text} are identical to the \modelname{RoBERTa} baselines) of this baseline. 

\noindent\textbf{SPQI Setup and Ablations.} We consider different model ablations using different feature subsets, and we distinguish between \modelname{SPQI-MoE} and \modelname{SPQI-CONCAT}, with the difference being  on how the diverse features are aggregated with \modelname{MoE} or simply concatenated. SPQI has four layers of graph convolutions~\cite{DBLP:conf/iclr/KipfW17}, and each node has 1024 dimensions.

\section{Evaluation Results}

\subsection{\revision{Offline Evaluation}}

\begin{table}[ht!]
    \centering
    \resizebox{1.0\linewidth}{!}{
    \begin{tabular}{ l | l l l | l l l | l l l | l l l | l l l | l l l }
    \toprule
    \textbf{Model} &  \multicolumn{3}{c}{Behavior}  &  \multicolumn{3}{c}{Product} &  \multicolumn{3}{c}{Text+Product}  &  \multicolumn{3}{c}{Text+Behavior} &  \multicolumn{3}{c}{Product+Behavior}  &  \multicolumn{3}{c}{Full} \\
    & \textbf{P} & \textbf{R} & \textbf{F1} & \textbf{P} & \textbf{R} & \textbf{F1} & \textbf{P} & \textbf{R} & \textbf{F1} & \textbf{P} & \textbf{R} & \textbf{F1} & \textbf{P} & \textbf{R} & \textbf{F1} & \textbf{P} & \textbf{R} & \textbf{F1}\\
    \midrule
         
         \modelname{MLP-Concat} & 0.827 & 0.910 & 0.866 & 0.758 & 0.755 & 0.757  & 0.758 & 0.766 & 0.762 & 0.856 & 0.916 & 0.885$^{*}$ & 0.876 & 0.881 & 0.878 & 0.875 & 0.881 & 0.878\\

        \modelname{MLP-MoE} &  0.907 & 0.886 & 0.896 & 0.756 & 0.758 & 0.757 & 0.761 & 0.730 & 0.744 & 0.895 & \textbf{0.921} & 0.907$^{*}$ & 0.903 & 0.908 & 0.906  & 0.903 & 0.914 & 0.908\\

        \modelname{SPQI-Concat} & 0.903 & 0.884 & 0.893$^{*}$ & 0.760 & 0.757 & 0.759 & 0.760 & 0.772 & 0.766 & 0.898 & 0.893 & 0.895 & 0.890 & 0.904 & 0.895$^{*}$  & 0.895 & 0.909 & 0.902$^{*}$\\
         
         \modelname{SPQI-MoE}  & 0.909 & 0.890 & 0.900$^{*}$ & 0.762 & 0.764 & 0.763$^{*}$ & 0.760 & 0.793 &  0.776$^{*}$ & \textbf{0.914} & 0.885 & 0.900 & 0.903 & 0.910 & 0.907 & 0.903 & 0.917 & \textbf{0.910}$^{*}$\\
    \bottomrule
    \end{tabular}}
    \caption{\small{SPQ detection performance of competing approaches. Significant results (as per paired t-test) are marked with $^*$ (\emph{p-value} $<0.05$). \modelname{SPQI-MoE} ablations are compared to  \modelname{MLP-MoE}, and \modelname{SPQI-Concat} vs. \modelname{MLP-Concat}. For the sake of clarity, we omit from the table the results for \modelname{RoBERTa-Question} and \modelname{RoBERTa-Text}, given that they are the lowest performing with F1=0.694 and F1=0.699, respectively. On the same feature set, \modelname{SPQI-MoE-Query} and \modelname{SPQI-MoE-Text} obtain significantly higher results with F1=0.777 and F1=0.774, respectively.}}
    \label{tab:performance}
\end{table}

\paragraph{Model Performance.} Table~\ref{tab:performance} shows the results of identifying SPQs. Our approach, \modelname{SPQI-MoE} achieves the best results with $F1=0.91$. \modelname{SPQI-Concat-Full} has a 1\% drop in F1. This shows that the flexibility of \modelname{Moe} to dynamically decide per question, which features are important in predicting user's shopping need. 

Comparing \modelname{SPQI-Full-MoE} against \modelname{MLP-MoE-Full}, the use of GATs provides an additional advantage in terms of F1. This shows that even for the same feature set and way to combine them, GATs ability to consider neighbouring nodes for determining shopping need is helpful. 
This in itself is intuitive and has been widely studied in recommender systems (e.g. collaborative filtering), where tying users with similar search patterns can be helpful in recommendations. In our case, according to how subgraphs are constructed for classification, our models exploit information coming from users who share similar purchasing patterns and ask similar questions.
Finally, \modelname{RoBERTa} baselines obtain the lowest performance, showing that identifying SPQs cannot be done from the question text alone.

In nearly all the cases, combining the question features using the \modelname{MoE} approach, allows the models to feature concatenation. Furthermore, computing the final question representation using GATs, where neighbouring nodes can have an impact on its representation, allows the model to obtain more accurate intent classification results.

Although the \modelname{SPQI} models are larger (in parameters) than their \modelname{MLP} counterparts, the performance decrease is not due to model size, as the comparison between \modelname{SPQI-MoE} and \modelname{SPQI-Concat} shows that \modelname{SPQI-MoE} achieves better performance.

\noindent\textbf{Ablations.}
The user behavioral features achieve the biggest performance improvement. \modelname{SPQI-Text+Behavior} is the most accurate model with $P=0.914$, which is 1\% higher than \modelname{SPQI-Full-MoE}. However, in terms of recall we get a drop of 3\%. The addition of product embedding features, allows the model to improve its recall by covering cases of users that may have a sparse purchase history or that few purchased products within the same category as the queried product. Interestingly, none of the feature combinations manage to achieve the full performance of \modelname{SPQI-MoE}, where we draw conclusions that while the precision may be higher on certain feature subsets (e.g. \modelname{SPQI-MoE-Text+Behavior}), the combination of all the features allows to obtain maximum coverage.

Similarly, here too, the \modelname{SPQ-MoE} and \modelname{SPQ-Concat} ablations perform better than most of their counterparts. In only one case the \modelname{MLP} obtain significantly better performance.

In terms of architecture, \modelname{SPQI-MoE-Text} obtains an increase of $F1=+7\%$ over \modelname{RoBERTa}. This improvement can be attributed to the use of GATs. Additionally, we do not see any significant difference between the models when trained only on the users questions, as opposed to both the question and the voice assistant answers. One reason may be that answers are usually handled by Q\&A systems that do not provide personalized answers to the users. 

Overall, only \modelname{MLP-MoE-Text+Behavior} achieves results that are significantly better than those of \modelname{SPQI-MoE\-Text+Behavior}. The performance decrease comes from lower recall.

In summary, the ablations show that our contributions are twofold. First, the proposed features that capture user's shopping need are highly suitable for voice search. Second, the proposed \modelname{MoE} for integrating the different features into a joint question representation, and the use of GATs for the intent classification task, with consistent improvement over the competing approaches (e.g., \modelname{MLP-Concat}/\modelname{MoE}). Note that the MLP baseline represents a strong baseline, since it is trained on the proposed features and uses the \modelname{MoE}.

\subsection{\revision{Online Recommendation Evaluation}}
\revision{
The SPQ definition in Section~\ref{sec:datasets} is only a proxy of shopping intent. PQs can theoretically be unrelated to the associated purchase. Our online evaluation, with real voice assistant users, verifies that the identified SPQs by our \modelname{SPQI-MoE} have a shopping intent. For this experiment, we split the users into two groups: 1) $T1$ where the users, according to our model, are recommended a shopping action to add a product into their shopping list, and 2) $C$ where all the users issuing a PQ are prompted to add the product into their shopping list. }

\revision{
$T1$ obtains 81.5\% higher F1 score than $C$ in identifying SPQs, and additionally users on $T1$  were 79.5\% more likely to add the product into their shopping list when compared to users in $C$. This experiment shows that our approach is able to identify user's shopping need, validated by users adding the queried product into their shopping list. Note that, not adding a product to the shopping list does not indicate the contrary, namely, the lack of shopping intent.}

%% file: conclusions.tex
\section{Conclusion}
\label{sec:conclusions}

We presented an approach to make proactive shopping recommendations in a leading industry voice assistant. By identifying shopping need, we allow the voice assistant to accurately recommend to its users relevant products or proactively suggest shopping actions that enhance their shopping experience.

We proposed a set of features that capture shopping intent from various perspectives, and relied on graph attention networks and on a novel mechanism to combine features for classifying a PQ as an SPQ. 
Experimental evaluations show that our proposed features and the way we encode user question through GATs, yield significant improvements over text classification approaches, achieving an increase in $F1$ of $+21\%$. The experiments confirm the highly latent nature of shopping intent.
This work, unlike others, explores users' shopping intent in voice, not only in the context of product search, but also in the context of product question answering. Furthermore, focusing on informational intent and exploring whether the underlying need behind is related to general knowledge or to shopping.